\documentclass[sigconf]{acmart}

\usepackage{xspace}
\usepackage[ruled]{algorithm2e}

\usepackage{array}

\newcommand{\repo}{https://github.com/ComplexData-MILA/misinfo-baselines}

\copyrightyear{2021}
\acmYear{2021}
\setcopyright{iw3c2w3}
\acmConference[WWW '21]{Proceedings of the Web Conference 2021}{April 19--23, 2021}{Ljubljana, Slovenia}
\acmBooktitle{Proceedings of the Web Conference 2021 (WWW '21), April 19--23, 2021, Ljubljana, Slovenia}
\acmPrice{}
\acmDOI{10.1145/3442381.3450111}
\acmISBN{978-1-4503-8312-7/21/04}

\makeatletter
\newcolumntype{^}{@{\hskip\tabcolsep\vrule width 1pt\hskip\tabcolsep}}
\makeatother
\usepackage{enumitem}

\AtBeginDocument{%
  \providecommand\BibTeX{{%
    \normalfont B\kern-0.5em{\scshape i\kern-0.25em b}\kern-0.8em\TeX}}}

\setcopyright{iw3c2w3}
\copyrightyear{2021}
\acmYear{2021}
\acmDOI{10.1145/3442381.3450111}

\acmConference[WWW ’21, April 19–23, 2021, Ljubljana, Slovenia]{WWW ’21, April 19–23, 2021, Ljubljana, Slovenia}{Apr 2021}{}
\acmBooktitle{The Web Conference '21}
\acmPrice{15.00}
\acmISBN{978-1-4503-XXXX-X/18/06}

\begin{document}


\title{The Surprising Performance of Simple Baselines for Misinformation Detection}


\newcounter{daggerfootnote}
\newcommand*{\daggerfootnote}[1]{%
    \setcounter{daggerfootnote}{\value{footnote}}%
    \renewcommand*{\thefootnote}{\fnsymbol{footnote}}%
    \footnote[2]{#1}%
    \setcounter{footnote}{\value{daggerfootnote}}%
    \renewcommand*{\thefootnote}{\arabic{footnote}}%
    }

\author{Kellin Pelrine}\authornote{Equal Contribution}
\email{kellin.pelrine@mila.quebec}
\affiliation{
  \institution{SOCS, McGill University}
  \institution{Mila - Quebec AI Institute}
}

\author{Jacob Danovitch}\authornotemark[1]
\email{jacob.danovitch@mila.quebec}
\affiliation{
  \institution{SOCS, McGill University}
  \institution{Mila - Quebec AI Institute}
}

\author{Reihaneh Rabbany}
\email{reihaneh.rabbany@mila.quebec}
\affiliation{
  \institution{SOCS, McGill University}
  \institution{Mila - Quebec AI Institute}
}




\renewcommand{\shortauthors}{Pelrine, Danovitch, Rabbany}


\begin{abstract}
As social media becomes increasingly prominent in our day to day lives, it is increasingly important to detect informative content and prevent the spread of disinformation and unverified rumours. While many sophisticated and successful models have been proposed in the literature, they are often compared with older NLP baselines such as SVMs, CNNs, and LSTMs. In this paper, we examine the performance of a broad set of modern transformer-based language models and show  that with basic fine-tuning, these models are competitive with and can even significantly  outperform recently proposed state-of-the-art methods.
We present our framework as a baseline for creating and evaluating new methods for misinformation detection. We further study a comprehensive set of benchmark datasets, and discuss potential  data leakage and the need for careful design of the experiments and understanding of datasets to account for confounding variables. As an extreme case example, we show that classifying only based on the first three digits of tweet ids, which contain information on the date, gives state-of-the-art performance on a commonly used benchmark dataset for fake news detection --Twitter16. We provide a simple tool to detect this problem and suggest steps to mitigate it in future datasets.

\end{abstract}




\maketitle

\section{Introduction}
\label{sec:intro}

Social media is filled with both treasure troves and landmines of information. There are far too many interactions to mine them by hand. But failing to understand them can have dire consequences. Misinformation can challenge fair elections \cite{MEEL2020112986} and cost billions of dollars \cite{rapoza2017can}. Misinformation spread in the COVID-19 "infodemic" \cite{cinelli2020covid, pulido2020covid} can cost lives. At the same time, successful mining of useful information can enable contact tracing and other measures to save lives \cite{9043580,Qin_2020}, as well as other applications like explainable fact checking \cite{cui2020deterrent}. 
Motivated by the profound impact, there has been substantial research on detecting fake news, misinformation, and related topics (such as bot and troll detection) in the past few years \cite{bondielli2019survey, survey2, mosinzova2019fake, survey4, covid19tweet}. A majority of these works are based on text/content classification \cite{shu2017fake,zhou2018fake,zhou2019fake,shu2019defend}. As real-world events such as the spread of COVID-19 misinformation show \cite{cinelli2020covid, pulido2020covid}, we are still far from addressing misinformation and more work is needed.

In this paper, we examine the use of modern pre-trained language models (LMs) for classifying content. Although language models are ubiquitous baselines in this domain, the ones used are often older models such as SVMs, CNNs, LSTMs, etc. \cite{denaux2020linked, VRoC20, dEFEND19, han2020graph}, instead of transformer-based models that have been dominant in NLP recently, such as the exemplar BERT model \cite{DBLP:journals/corr/abs-1810-04805, xia2020bert}. Here, we report the performance of these recently proposed language models on common benchmark datasets for misinformation detection. More specifically, we consider small, medium and large models and show that most can achieve comparable or even better performance than state-of-the-art (SOTA) methods. Even the smallest 4 million parameter BERT-Tiny \cite{turc2019} can approach or beat state-of-the-art in some cases. These SOTA methods are usually much more complex and in many cases incorporate more information beyond the content, e.g. the reply thread. Although some of these more sophisticated methods are also designed to incorporate language models, it is not straightforward how to change their language module and they may or may not synergize well with the latest language models. 

We further study commonly used benchmark datasets and report a significant issue with the Twitter15 and Twitter16 datasets \cite{liu2015real, ma2016detecting, ma2017detect}, which can produce unrealistically high performance just by inferring the tweet date. This issue is also present to a lesser degree in FakeNewsNet \cite{shu2018fakenewsnet}. We propose a simple test to identify it in other datasets and suggest how to mitigate it in data collection. 

We also provide the exact splits to use our results as benchmarks, as well as other suggestions on proper evaluation techniques, and discuss potential future directions of work in this domain.

To summarize, the main contributions of this paper are threefold:
\begin{itemize}[noitemsep,topsep=0pt,leftmargin=*]
    \item We show that recent pretrained LMs are competitive with and sometimes even substantially better than SOTA models in common benchmarks for misinformation detection.

    \item We discuss experiment design considerations, effect of different data splits, and further demonstrate potential data leakage issues on commonly used benchmark datasets.

    \item We package our framework as a simple yet strong baseline for relevant tasks, and provide the tweet IDs to exactly reconstruct our splits, facilitating use of our reported results for benchmarks. Our framework is available on \href{\repo}{GitHub}. 
\end{itemize}

\section{Background and Related Work}
\label{sec:background}

There are two key literatures related to this work: (mis-)information classification on the application domain side, and natural language processing models on the methodological side. We discuss first the former and then the latter.

\subsection{Detecting Misinformation}
We can broadly divide misinformation detection algorithms into methods that analyze content and (social) context. Here, we focus on the methods that are content-based. For more complete review of the related works, please refer to the recent surveys \cite{bondielli2019survey, survey2, mosinzova2019fake, survey4,zhou2019fake,shu2019defend}. 
The content has been analyzed previously on multiple levels. At the word level, methods have used features including word counts, derived measures like TF-IDF, small combinations of words like bigrams, and word embeddings \cite{WWWSyntax20, rubin2016fake, castillo2011information, benamira2019semi}. At the sentence level, methods rely on features like syntax and complexity \cite{WWWSyntax20, rubin2016fake, castillo2011information, benamira2019semi}. Finally, there are features derived from the text as a whole, including general representations and more specialized ones such as topic or sentiment analysis \cite{VRoC20, SAME19, RumorSleuth19, ito2015assessment}.

More specifically, we consider the following state-of-the-art algorithms as a representative set of the current methods: 

\begin{itemize}
    \item \citet{VRoC20} use a multitask (rumor detection, tracking, veracity classification, and stance classification) LSTM-based VAE \cite{kingma2013auto}, with the text as input. 
    \item \citet{han2020graph} use a GNN \cite{scarselli2008graph} to encode the propagation network and features of the users in it (not including text).
    \item \citet{huang2020heterogeneous} build a heterogeneous tweet-word-user graph, linking the representations with an attention mechanism.
    \item \citet{lu2020gcan} use a co-attention mechanism to combine text and propagation information.
    \item \citet{dEFEND19} use attention mechanisms to combine word- and sentence-level features from articles, encoded with a GRU, with a comment (tweet) encoder.
    \item \citet{wang2020fake} link text and object detections in images with entities and a knowledge graph constructed from them.
    \item \citet{wu2020dtca} uses a combination of a decision tree model and co-attention to evaluate credibility of tweets using text, propagation, and user information.
    \item \citet{wu2020rumor} build a propagation graph from tweets and replies. They embed each using Doc2Vec \cite{le2014distributed}, and use a GRU (Gated Recurrent Unit) \cite{cho2014learning} and attention mechanism to pass information and pool the graph.
\end{itemize}

The most related work to ours is \cite{huang2020conquering}. They highlight "cross-source" failure of existing misinformation detection methods, in terms of training on one dataset and testing on another, and also propose a new model designed to solve it. Our results support and expand on theirs by identifying "cross-topic" and "cross-domain" effects even within the same datasets, depending on how they are split, as well as differences between datasets. We take a different direction for modeling: rather than focusing on cross-source tasks and developing a specialized model, we look at a spectrum of tasks and ask how general language models perform in this domain.

\subsection{Detecting Informative Content}

Instead of detecting which tweets are spreading misinformation, there are also many contexts in which we want detect tweets that provide useful information. Recently, there has been a big focus in the research community on COVID-19 \cite{brainard2020scientists}. Mining information about disaster events is not new \cite{imran2016lrec, takahashi2015communicating, kankanamge2020determining, chen2020uncovering, kumar2011tweettracker, to2017identifying, singh2019event, aggarwal2019classification}. But COVID-19 is unprecedented in its global scale and impact, and requires new methods and understanding.

Although the focus of our experiments is on misinformation, we include a dataset for detecting informative COVID-19 tweets - discussed in more detail in section 3.2 - to highlight the connections between these two areas and how language models can model both as content classification. 

\subsection{Pre-Trained Language Models}

The invention of word embeddings was a fundamental breakthrough \cite{mikolov2013efficient,mikolov2013distributed} leading towards today's language models. By predicting words, one can go beyond one-hot and similar encodings and convert words into vectors that encode their meaning and are more useful for downstream tasks. 

The first embedding models such as Word2Vec \cite{mikolov2013efficient,mikolov2013distributed}, GloVe \cite{pennington2014glove}, and FastText \cite{joulin2016bag} generate fixed vectors for each input word or token. This has the advantage of simplicity, but can struggle to capture words whose meanings vary depending on the context, and in turn struggle with downstream tasks that require this level of language understanding. Therefore, better methods to incorporate context became a key research topic. A followup direction was to use convolutional neural networks, ubiquitous in computer vision, on text \cite{kim2014convolutional}. These could effectively capture local context, but still missed incorporating longer range context. Similarly, RNNs, which could capture local sequences, had problems with vanishing gradients and could not capture longer range information \cite{minaee2020deep}.

LSTMs \cite{hochreiter1997long}, which like RNNs operate on sequences but allow longer range dependencies, helped solve this issue. There are state of the art models for misinformation detection, including the ones discussed in the domain section above, which use various versions or improvements on LSTMs, such as the GRU by \citet{cho2014learning}. These also led to contextualized embeddings such as ELMo \cite{Peters:2018}.

Another major breakthrough came with attention \cite{bahdanau2014neural,yang2016hierarchical} and then, based on attention, transformers \cite{vaswani2017attention}. These facilitate far more parallelization than LSTMs \cite{minaee2020deep}, which means models based on them can be trained on much larger and more general corpora. A key early model using these is BERT \cite{DBLP:journals/corr/abs-1810-04805} (discussed further in the following section), and there has subsequently been numerous different models proposed based on transformers \cite{xia2020bert}. These vary from small tweaks to existing models such as pretraining on domain-specific data \cite{covid-twitter-bert, BERTweet}, to significantly different models such as GPT (and then GPT-2 and GPT-3) \cite{radford2018improving, radford2019language, brown2020language}. Virtually all of these models can provide contextualized word/sentence/document embeddings for downstream tasks. These models are easily accessible through packages such as \cite{Wolf2019HuggingFacesTS}, and with hundreds of versions to choose from, as \citet{xia2020bert} suggests the hardest question may be which one to use.

Although there are many excellent options, a large number of recent papers in the misinformation domain still benchmark their methods against older models \cite{denaux2020linked, VRoC20, dEFEND19, han2020graph, cui2020deterrent, wu2019different, ajao2019sentiment}. While there are still contexts in which these models do well and there is nothing wrong with making these comparisons, our results show that this can sometimes result in weak benchmarks when the more modern language models are omitted. Among other results, we give suggestions on how to incorporate these new models into evaluation in this domain. Please note that although training these newest models from scratch can require a prohibitive amount of computational resources, pretrained weights are readily available and are commonly used instead. 

\section{Methodology}
\label{sec:exp}

\newcommand{\cls}{\texttt{[CLS]}}
\newcommand{\modeldescription}[4]{\vspace{2pt}
\textbf{#1} (\texttt{#2}) \cite{#3}. \\#4
}
Our baselines simply consist of a set of language models and benchmark datasets. 
Here we first explain the language models we consider in our baseline framework in Section \ref{sec:models}, then we describe the benchmark datasets in Section \ref{sec:data}. For the Funnel Transformer and ALBERT, we employ a mean pooling strategy over the final layer of output. For ELMo we use an LSTM pooler. For all others, we use the \cls\: token embedding from the final layer of output. For all models, we allow all parameters of the language model to be fine-tuned, use a single fully-connected layer on the pooled embeddings for classification, train using cross-entropy loss, and use \texttt{AdamW} \cite{loshchilov2017decoupled} with a slanted triangular learning rate scheduler for optimization.

\subsection{Language Models}
\label{sec:models}

We evaluate a variety of language models. Where available, we use the implementations and pre-trained weights provided by the Huggingface Transformers library \cite{Wolf2019HuggingFacesTS}, and train using PyTorch \cite{Paszke2017AutomaticDI} and AllenNLP \cite{Gardner2017AllenNLP}. We include the following language models (with the exact version\footnote{As available at \url{https://huggingface.co/models}} in parentheses):

\modeldescription{BERT}{bert-base-uncased}{DBLP:journals/corr/abs-1810-04805}{Bidirectional Encoder Representations from Transformers, or BERT for short, is a large pre-trained bidirectional transformer. BERT was pre-trained using two objectives. The first, masked language modelling, required BERT to predict a masked token from the input. The second, next sentence prediction, required predicting whether two sentences appeared consecutively in the training corpus. To complete the latter task, BERT prepends the input text with a special \cls\: token, and inserts a special \texttt{[SEP]} token between the first and second sentence as well as at the end of the second sentence. The \cls\: token is commonly used as a document-level representation for classification tasks.
}

\modeldescription{BERT-Tiny}{bert\_uncased\_L-2\_H-128\_A-2}{turc2019}{
BERT-Tiny is the smallest of several pre-trained language models that follow the same pre-training procedure as BERT, as well as a knowledge distillation fine-tuning procedure. These smaller models achieve strong performance on downstream tasks with significantly fewer parameters. 
}

\modeldescription{RoBERTa}{roberta-large} {Liu2019RoBERTaAR}{RoBERTa follows a similar architecture to BERT but removes the next-sentence prediction pre-training objective while making the masking procedure dynamic by regenerating the mask for each example every time and leverage larger batch sizes and increased training iterations to improve performance. They find that BERT underfits its training data. }

\modeldescription{ALBERT}{albert-large-v2} {Lan2020ALBERTAL}
{ALBERT adds an additional pre-training objective while incorporating two methods that reduce the number of parameters in the model, factorizing the embedding layer and tying weights across hidden layers.} 

\modeldescription{BERTweet} {bertweet-base}{BERTweet}{
BERTweet follows an identical training procedure to RoBERTa, and is pre-trained on 850 million tweets.
}

\modeldescription{COVID-Twitter-BERT} {covid-twitter-bert-v2}{covid-twitter-bert}{
COVID-Twitter-BERT (CT-BERT) follows an identical training procedure to BERT, and the latest version is pre-trained on 1.2 billion training examples generated from 97 million tweets.}

\modeldescription{DeCLUTR} {declutr-base} {Giorgi2020DeCLUTRDC}{DeCLUTR is a transformer-based language model that proposes a contrastive, self-supervised method for learning general purpose sentence embeddings. The model is trained with a masked language modelling objective as well as with contrastive loss using both easy and hard negatives.}

\modeldescription{Funnel Transformer} {xlarge-base} {dai2020funneltransformer}{
The Funnel Transformer improves the efficiency of bidirectional transformer models by applying a pooling operation after each layer, akin to convolutional neural networks, to reduce the length of the input.
}

\modeldescription{ELMo} {Original\footnote{Available at \url{https://allennlp.org/elmo}}} {Peters:2018}{
ELMo is one of the first large-scale pre-trained language models. It is a character-based model and is the only model in this list that does not optimize for masked language modelling. Instead, it uses bidirectional LSTMs \cite{hochreiter1997long} to perform autoregressive language modelling in both the forward and backward directions. 
}

\subsection{Benchmark Datasets}
\label{sec:data}
We consider the comprehensive set of benchmark datasets available in the literature for misinformation (information) detection. In the following we briefly explain each dataset.

\subsubsection{PHEME}
The PHEME dataset \cite{zubiaga2016analysing, kochkina2018all} contains 6425 tweets about 9 newsworthy events. The events are unrelated. The tweets were collected as the stories developed, with a journalist annotating them as rumour vs. non-rumour. They grouped them by story, and then marked whether the stories were true or false once it was confirmed, or as unverified if they could not be certain during the collection period. There are also other labels such as stance, which we do not consider here.

\newcommand{\datadescription}[4]{\vspace{2pt}
\textbf{#1 #2} which matches \citet{#3}. \\#4
}
We split this dataset five different ways to compare with different state-of-the-art results that use the corresponding settings. 

\datadescription{PHEME9}{T/F}{wu2020adaptive}{
First, we take only the tweets marked as true or false and do a 70-10-20 train-dev-test split. }

\datadescription{PHEME5}{R/NR}{wang2020fake}{
Next we take the 5 largest events (comprising ~90\% of the dataset) and use the rumour and non-rumour labels. We again split 70-10-20. This setup matches both \cite{wang2020fake} (70-30 split with no dev set) and \cite{cheng2020vroc} (which notes 10\% withheld for validation). }

\datadescription{PHEME5}{3-way}{cheng2020vroc}{
Here, we take the 5 largest events and tweets labeled true, false, and unverified, splitting 70-10-20. }

\datadescription{PHEME9}{4-way}{wu2020rumor}{
For this, we subsample 800, 400, 600, 500 tweets with at least 3 replies, that are non-rumor, false, true, and unverified, respectively. They are split 80-10-10. }

\datadescription{PHEME5}{Lc}{cheng2020vroc}{
Finally, we also examine an event-based split. Starting with the 5 largest events, we train on four and test on the largest (tweets related to Charlie Hebdo terror attack), with the 3-way true, false, and unverified labels.
}

\subsubsection{\textbf{FakeNewsNet}}

The FakeNewsNet dataset \cite{shu2018fakenewsnet, shu2017exploiting, shu2017fake} contains articles fact-checked by PolitiFact\footnote{https://www.politifact.com/} or GossipCop\footnote{https://www.gossipcop.com/}, and related tweets. The labels are "real" and "fake." We retrieve 467 thousand tweets in the \textbf{PolitiFact} dataset, and 1.25 million in \textbf{GossipCop}. We split this dataset by first splitting the articles, then assigning tweets to each split based on the article they are associated with. We divide it 75-10-15, corresponding to the splits in \cite{shu2019defend, han2020graph, shu2020leveraging}. We remove two events from Politifact that are labeled both real and fake, politifact14920 and politifact14940.
Work on this dataset such as \cite{shu2019defend, han2020graph} has focused on classifying the articles, and using the tweets and related user information as supplemental information to improve performance. Because our pipeline was set up for tweet data rather than articles, we apply a simple way of classifying the articles through the tweets, by classifying the tweets and converting the predictions to an article prediction based on majority vote of the corresponding tweets' labels. Note that the works we compare with use the tweets, so this is not using an additional type of data that they have excluded. Following \cite{shu2019defend, han2020graph}, we evaluate only on articles with at least 3 corresponding tweets. 

\subsubsection{\textbf{Twitter15} and \textbf{Twitter16}}

These datasets \cite{liu2015real, ma2016detecting, ma2017detect} contain tweets related to stories from fact-checking websites and random tweets. There are 1490 and 818 tweets respectively, labeled true, false, unverified, or non-rumor. We reserve 10\% for the dev set, then split the remainder 75:25, matching \cite{huang2020heterogeneous}.

\subsubsection{\textbf{Twitter15 T/F} and \textbf{Twitter16 T/F}}
These are Twitter15 and Twitter16 with the true and false examples only. This gives 742 and 412 examples respectively. We split 70-10-20, giving the same training set size as \cite{lu2020gcan}, which split 70-30 with no dev set. 

\subsubsection{\textbf{WNUT-2020}}

This dataset was created for a shared task in WNUT-2020 on classifying tweets as ``informative'' or ``uninformative'', in providing information about "recovered, suspected, confirmed and death cases as well as location or travel history of the cases" \cite{covid19tweet}. Many of the models submitted in the competition were based on pre-trained transformer-based language models \cite{wadhawan2020phonemer, tran2020uit,Nguyen2020TATLAW,Huynh2020BANANAAW,Nguyen2020TATLAW,Huynh2020BANANAAW,Babu2020CIANITTAW,Perrio2020CXP949AW,Maveli2020EdinburghNLPAW,Chauhan2020NEUAW}, including the first place model which was an ensemble of CT-BERT and RoBERTa \cite{kumar2020nutcracker}, and the tied first place model which was a carefully tuned version of CT-BERT \cite{moller2020nlp}. A 70-10-20 split was provided by the task organizers.

\subsubsection{\textbf{CoAID}}

The CoAID dataset \cite{cui2020coaid} is a misinformation dataset related to COVID-19. This dataset is new and consists of tweets, articles, and claims. The authors provide baselines on the articles. To the best of our knowledge, as of this writing, no existing work has performed classification on the tweets. 

We conducted experiments here aimed to establish a baseline for future works. However, we found that most splits result in virtually all models obtaining near-perfect (above 95) F1 score. We examined in particular the ``news'' tweets. Many tweets about the same news item have similar or even duplicated text, so we tried splitting similar to FakeNewsNet, first splitting the news articles and then assigning tweets to each set based on their corresponding article. We also tried subsampling a small train set to make the problem harder. This did result in one set of results with lower F1, but we found it was unstable and did not replicate consistently when we redid the splitting and training process. Thus, we do not report results on this dataset here. However, it contains a significant amount of useful data on an important problem, so we encourage future work to determine challenging splits and help others take full advantage of this resource.

\subsection{Implementation Details}

 Each model is trained on an RTX8000 GPU using mixed precision with a batch size of 32. and learning rate of $1\mathrm{e}$-5. We do not perform any hyperparameter tuning. We train for 50 epochs on all datasets except Politifact and Gossipcop. There we train for 2 epochs, because they are much larger than the rest. 
 
 We run each model to completion exactly 5 times, and report mean and standard deviation, with two exceptions. First, datasets with a fixed split (PHEME5 Lc and WNUT-2020), which are run once. Second, a bug caused a small number of GossipCop results with Funnel to be lost, and due to time constraints we were unable to fully repeat the experiments. Thus we do not report results for that model and dataset.
 
 Complete implementation details, hyperparameter configurations, and tweet IDs for each split of each dataset can be found on GitHub.\footnote{\repo}

\subsection{Experimental Results}
\label{sec:res}
\subsubsection{Performance Evaluation}

We roughly categorize the algorithms by size. The first set are "large" models with around 400 million parameters. The second set are "medium" ones with around 100 million. The third set are "small" ones under 20 million. Order in the tables is alphabetical within category. The exact parameter counts are shown in table~\ref{tab:num_params}.

Results are shown in tables \ref{tab:PHEME_results} and \ref{tab:other_results}. The first row presents the algorithm that achieves, to the best of our knowledge, state-of-the-art (SOTA) performance on the given dataset and split.

\begin{table*}[htbp]
\centering
\caption{Number of Parameters}
\label{tab:num_params}
\begin{tabular}{|c|c|c|c|c|c|c|c|c|}
\hline  
CT-BERT & Funnel & RoBERTa & BERT & BERTweet & DeCLUTR & ELMo & ALBERT & BERT-Tiny \\ 
\hline  
340M & 468M & 355M & 110M & 135M & 125M & 94M & 17M & 4M  \\ 
\hline
\end{tabular} 
\end{table*}

\begin{table*}
\centering
\caption{PHEME Results (Macro F1 score)}
\label{tab:PHEME_results}
\begin{tabular}{|c|c|c|c|c|c^c|}
\hline  
& PHEME9 T/F & PHEME5 R/NR & PHEME5 3-way & PHEME9 4-way & PHEME5 Lc & Average Rank \\ 
\hline  
SOTA & 82.5 \cite{wu2020dtca} &  87.6 \cite{cheng2020vroc, wang2020fake} & 66.7 \cite{cheng2020vroc} & 75.3 \cite{wu2020rumor} & \textbf{51.3} \cite{cheng2020vroc} & 5.6 \\ 
\hline 
\hline
CT-BERT & 92.0 $\pm$ 0.9 & 89.0 $\pm$ 0.8 & 84.6 $\pm$ 1.5 & 79.0 $\pm$ 2.6 & 27.9 & 3.4 \\ 
\hline 
Funnel & 86.7 $\pm$ 3.2 & 87.3 $\pm$ 0.6 & 79.4 $\pm$ 3.7 & 71.4 $\pm$ 3.3 & 28.7 & 6.4 \\
\hline 
RoBERTa & \textbf{93.2 $\pm$ 0.9} & \textbf{89.4 $\pm$ 0.3} & \textbf{87.7 $\pm$ 1.9} & \textbf{82.5 $\pm$ 3.3} & 29.0 & \textbf{2.0} \\ 
\hline 
\hline 
BERT & 89.9 $\pm$ 1.1 & 87.2 $\pm$ 0.4 & 81.2 $\pm$ 1.4 & 76.8 $\pm$ 2.7 & 24.2 & 5.8 \\ 
\hline 
BERTweet & 89.8 $\pm$ 0.6 & 87.3 $\pm$ 0.6 & 81.8 $\pm$ 0.9 & 76.6 $\pm$ 4.1 & 29.0 & 5.0\\ 
\hline 
DeCLUTR & 90.2 $\pm$ 0.8 & 88.3 $\pm$ 0.4 & 83.7 $\pm$ 2.1 & 77.8 $\pm$ 3.5 & 30.2 & 3.2 \\ 
\hline 
ELMo & 81.7 $\pm$ 2.4 & 84.2 $\pm$ 0.8 & 65.8 $\pm$ 1.8 & 64.3 $\pm$ 4.0 & 30.3 & 9.4 \\ 
\hline
\hline 
ALBERT & 85.3 $\pm$ 2.9 & 84.2 $\pm$ 2.7 & 71.1 $\pm$ 2.2 & 65.7 $\pm$ 3.1 & 29.4 & 7.2 \\ 
\hline  
BERT-Tiny & 81.6 $\pm$ 2.0 & 84.7 $\pm$ 0.8 & 67.3 $\pm$ 2.0 & 61.0 $\pm$ 2.5 & 36.5 & 7.6 \\ 
\hline
\end{tabular} 
\end{table*}

\newcommand{\declutrfootnote}{\footnote{Although in most cases our untuned hyperparameters work well, it appears they are not appropriate for DeCLUTR on FakeNewsNet.}}

\newcommand{\bugnote}{\footnote{Due to a bug, runs for this model and dataset were lost. See section 3.3 for details.}}

\newcommand{\otherresults}{
\hline  
& PolitiFact & GossipCop & Twitter15 & Twitter16 & Twitter15 T/F & Twitter16 T/F & WNUT-2020 & Average Rank \\ 
\hline  
SOTA & \textbf{92.8} \cite{shu2019defend} & 85.0 \cite{han2020graph} & \textbf{91.0} \cite{huang2020heterogeneous} & \textbf{92.4} \cite{huang2020heterogeneous} & 82.5 \cite{lu2020gcan} & 75.9 \cite{lu2020gcan} &  \textbf{91.0} \cite{kumar2011tweettracker, moller2020nlp} & 4.4 \\ 
\hline 
\hline
CT-BERT & 86.0 $\pm$ 3.2 & 90.6 $\pm$ 0.2 & 83.5 $\pm$ 2.8 & 83.9 $\pm$ 0.9 & 93.8 $\pm$ 1.6 & 94.0 $\pm$ 3.5 & 90.6 & 2.8 \\ 
\hline 
Funnel  & 86.4 $\pm$ 3.2 & --\bugnote & 66.9 $\pm$ 3.0 & 69.6 $\pm$ 2.9 & 83.2 $\pm$ 3.8 & 90.8 $\pm$ 2.2 & 88.5 & -- \\ 
\hline 
RoBERTa & 86.7 $\pm$ 1.2 & \textbf{92.8 $\pm$ 0.5} & 81.8 $\pm$ 1.5 & 84.8 $\pm$ 1.9 & \textbf{94.4 $\pm$ 0.8} & \textbf{95.7 $\pm$ 2.8}  & 90.5 & \textbf{2.3}  \\ 
\hline  
\hline 
BERT & 81.8 $\pm$ 3.0 & 89.8 $\pm$ 0.4 & 77.5 $\pm$ 3.3 & 78.2 $\pm$ 4.1 & 89.7 $\pm$ 1.6 & 91.6 $\pm$ 4.5 &  88.5 & 5.3 \\ 
\hline 
BERTweet & 88.5 $\pm$ 1.2 & 92.6 $\pm$ 0.6 & 76.7 $\pm$ 2.9 & 77.7 $\pm$ 2.7 & 86.7 $\pm$ 1.8 & 92.0 $\pm$ 3.7 & 88.8 & 4.4 \\ 
\hline 
DeCLUTR & 36.6 $\pm$ 1.4\declutrfootnote & 43.3 $\pm$ 0.4\textsuperscript{b} & 80.4 $\pm$ 2.6 & 80.5 $\pm$ 1.7 & 91.7 $\pm$ 1.5 & 94.5 $\pm$ 2.5  & 89.1 & 5.1 \\ 
\hline 
ELMo & 83.1 $\pm$ 1.6 & 92.0 $\pm$ 0.5 & 53.7 $\pm$ 2.7 & 55.5 $\pm$ 4.9 & 74.4 $\pm$ 3.5 & 83.3 $\pm$ 5.0 & 82.4 & 8.0 \\ 
\hline 
\hline 
ALBERT & 80.1 $\pm$ 2.9 & 88.2 $\pm$ 0.9 & 63.4 $\pm$ 4.0 & 68.0 $\pm$ 3.5 & 83.3 $\pm$ 1.8 & 88.9 $\pm$ 4.3  & 86.8 & 7.7 \\ 
\hline 
BERT-tiny & 85.3 $\pm$ 2.8 & 86.5 $\pm$ 0.6 & 54.6 $\pm$ 3.4 & 48.8 $\pm$ 3.9 & 77.8 $\pm$ 4.8 & 77.8 $\pm$ 4.9 &  79.9 & 8.9  \\  
\hline 
}

\begin{table*}
\begin{minipage}{\textwidth}
\centering
\small
\caption{Other Dataset Results (Macro F1 score)}
\label{tab:other_results}
\begin{tabular}{|c|c|c|c|c|c|c|c^c|}
\otherresults
\end{tabular} 
\end{minipage}
\end{table*}





We discuss these results in section 4.

\subsubsection{Potential Data Leakage}

During our analysis, we observed a trend in the Twitter15 and Twitter16 dataset. It appears that time is a highly discriminative factor in separating the classes. This creates the possibility of data leakage by means of several potential confounding variables. To illustrate, we demonstrate that competitive performance can be achieved by using only the first 2 or 3 digits of the tweet ID. Due to the way unique IDs are generated for tweets, these digits reveal information about the time the tweet was posted.\footnote{\url{https://github.com/client9/snowflake2time}} We train a random forest on these digits, setting a high depth (25, such that increasing it further does not change validation performance) and otherwise using the default \texttt{scikit-learn} \cite{scikit-learn} parameters.

The class labels are balanced, so a random baseline will have 25\% F1. This is roughly what one would intuitively expect from a model using only IDs as features, because the time, particularly at this approximate scale, should have little correlation if any with whether a tweet is true or false, and would not be useful on its own for classification in applied settings. Unfortunately, as shown in table \ref{tab:tweet_id_results}, this is not the case. The IDs perform moderately well on Twitter15, and approach state-of-the-art on Twitter16. The non-rumor class on Twitter15 is also trivial to detect from the date.

\begin{table*}[htbp]
\begin{center}
\caption{Evaluating tweet ID classification (Macro F1 score)}
\label{tab:other_tweet_id_results}
\begin{tabular}{|c|c|c|c|c|c|}
\hline  
\textbf{Twitter15} & False & True & Unverified & Non-rumor & Macro Avg.\\ 
\hline 
SOTA \cite{huang2020heterogeneous} & 92.9 & 90.5 & 85.4 & 95.3 & 91.0 \\
\hline 
2-digit RF & 62.4 & 65.6 & 61.1 & 99.4 & 72.1 \\ 
\hline 
3-digit RF & 73.0 & 69.5 & 79.7 & 98.2 & 80.1 \\ 
\hline 
\hline
\textbf{Twitter16} & False & True & Unverified & Non-rumor & Macro Avg.\\
\hline
SOTA \cite{huang2020heterogeneous} & 91.3 & 94.7 & 89.9 & 93.5 & 92.4 \\
\hline 
2-digit RF & 83.5 & 87.6 & 82.1 & 90.7 & 86.0 \\ 
\hline 
3-digit RF & 90.7 & 95.3 & 84.4 & 92.9 & 90.8 \\ 
\hline
\end{tabular} 
\end{center}
\end{table*}

Of course, classification using tweet IDs alone is uninteresting in isolation. However, this paints a broader picture of potential confounding variables within these datasets (and others collected in the same manner). Tweet IDs are not the only things that evolve over time; the content posted to Twitter itself changes drastically over time, as language evolves and topics enter and exit public discourse. For example, consider two particularly time-sensitive words: "Clinton" and "Trump", the surnames of the 2016 US presidential candidates. The presence of these words, like the tweet IDs, is revealing about the time at which a tweet was posted (and therefore the potential label distribution), as public discourse largely focused on their campaigns throughout 2016. In table \ref{tab:tweet_id_results}, we see that the presence of those words (uncased) alone allow us to rule out "true," and in the case of Twitter15 also "false." This is obviously very unrealistic, and can cause a classifier to rely too much on particular keywords and fail to generalize to unseen data in applied settings.

\begin{table*}[htbp]
\begin{center}
\caption{Label counts of tweets containing "Clinton" and "Trump"}
\label{tab:trump_clinton_results}
\begin{tabular}{|c|c|c||c|c|c|}
\hline  
\textbf{Twitter15} & Clinton & Trump & \textbf{Twitter16} & Clinton & Trump \\ 
\hline  
True & 0 & 0 & True & 0 & 0\\ 
\hline 
False & 0 & 0 & False & 17 & 18\\ 
\hline 
Unverified & 22 & 30 & Unverified & 17 & 39\\ 
\hline 
Non-rumor & 6 & 14 & Non-rumor & 8 & 6\\ 
\hline 
\end{tabular} 
\end{center}
\end{table*}

After discovering this problem with Twitter15 and Twitter16, we next examined the other datasets. Results are shown in table~\ref{tab:other_tweet_id_results}. Here \textbf{Top Model} refers to the best model reported in this paper, either the SOTA model (indicated by a citation) or one of our language models. \textbf{PHEME9 All} is comprised of all PHEME tweets with the full 4-way labels. We did not compare results on this split, so we do not list a top model entry in that column. \textbf{Random} refers to a stratified random classifier, i.e. outputting a class randomly with probability according to the training label distribution.

\begin{table*}[htbp]
\begin{center}
\caption{Evaluating tweet ID classification (F1 score)}
\label{tab:tweet_id_results}
\begin{tabular}{|c|c|c|c|c|c|}
\hline  
& PolitiFact & GossipCop & PHEME9 All & CoAID & WNUT-2020 \\ 
\hline 
Top Model & 92.8 \cite{shu2019defend} & 91.0 & - & 82.2 & 91.6 \\
\hline 
Random & 50.1 & 49.9 & 26.8 & 35.1 & 51.4 \\
\hline
2-digit RF & 77.3 & 65.8 &  25.6 & 49.5 & 34.6\\ 
\hline 
3-digit RF & 76.2 & 66.2 & 43.5 & 46.2 & 51.4 \\ 
\hline 
\end{tabular} 
\end{center}
\end{table*}

The issue is not as pronounced on these datasets as on Twitter15 and Twitter16, but is still present, especially in PolitiFact and GossipCop. CoAID and PHEME are not too bad; classifying the IDs gives better than random performance, but the performance is still bad. WNUT-2020 is split particularly well in time, as classifying the IDs is no better than random.

\begin{figure*}
  \centering
    \caption{False vs. Other Classes}
    \label{fig:false_v_other}
    \Description{Scatter plot of words and labels}
  \includegraphics[width=\linewidth]{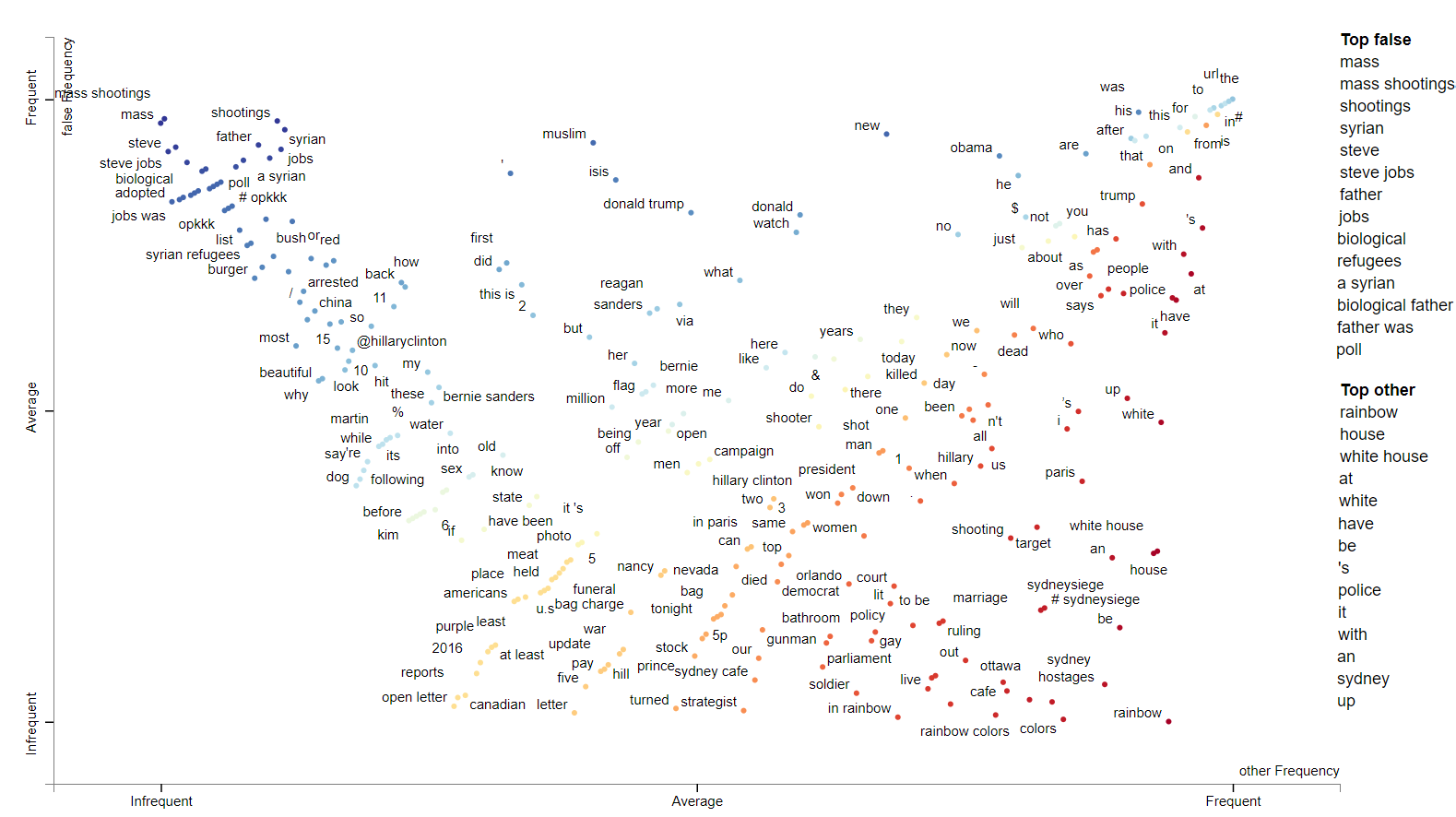}
\end{figure*}

To examine the data from another perspective, we can look at words by label in Twitter16. In figure~\ref{fig:false_v_other}, we show an example visualization of the label "false" versus the rest. We can see that "steve" is extremely correlated with a label of "false." This is due to a false tweet about Steve Jobs being adopted that is exactly duplicated 12 times, and nearly duplicated an additional 5. In the real world, it is unlikely that "Steve" guarantees a false label, but in this data it does. We can also see that there seem to be some false tweets about mass shootings - again, unlikely to be so strongly discriminative in the real world.

\section{Discussion}
\label{sec:concl}

These datasets fall on a spectrum of required generalization. On one end, there are "in-topic" datasets such as WNUT-2020 and the four PHEME splits not including PHEME5 Lc. These correspond to a real-world problem where one wants to detect misinformation or other content on a known topic. For example, a current task is detecting 5G COVID-19 conspiracy tweets \cite{mediaeval2020fake5g}. On the other end, there are "cross-domain" datasets, such as PHEME5 Lc, where although the task is shared, the content of the test examples is otherwise unrelated to the train examples. A real-world application of this is detecting misinformation about a new event. In the middle, there are datasets whose examples share related domains, but have somewhat different topics, such as PolitiFact.

The language models examined here excel at "in-topic" classification, often beating state-of-the-art methods by large margins in the results here. On the other hand, they perform poorly at cross-domain tasks. For example, they produce very mediocre performance on PHEME Lc, likely due to overfitting to the topics of the training set and failing to generalize to a different topic. State-of-the-art models in this setting, such as \cite{cheng2020vroc}, use other information beyond content to avoid this failure. Meanwhile, in between the two extreme types of datasets, we see more variable results. For example, LMs beat state-of-the-art on GossipCop but not PolitiFact.

Real-world problems can fall anywhere along the entire spectrum, and even the best results on these datasets - whether ours or others' - still fail on a significant fraction of detections. So better methods are needed on all counts. The results here suggest that understanding where one's task falls on this spectrum is very helpful. When it is further towards the "in-topic" side, recent language models may be very effective, at minimum providing strong benchmarks, and potentially a foundation on which to build more sophisticated models.

Comparing new domain-specific models with older language models may complicate interpretability of the results. The obvious, definitive solution is to always compare with the latest language models. However, due to time and computation constraints, this may not always be feasible - for example, even the largest models we report here are not the largest models available. There are three alternatives. One is to match splits and rely on other papers that run the latest LMs. This is may not be viable for new data, but we encourage work on standard datasets to provide solid benchmarks, and matching splits with those benchmarks whenever possible to facilitate comparison. As noted in the introduction, we provide IDs for our exact splits so that our results can be used in this way.\footnote{\repo} Another alternative is to build models which can incorporate the latest language model, and provide evidence that it improves results compared to language models alone (and does so better in some way than existing work). These may be particularly valuable, as they are to some degree future-proofed, or at least future-adaptive, since they can be updated with newer language models. Finally, researchers can explore different directions that are orthogonal to content, and give evidence that they provide information that cannot be extracted from the content by language models. There is already a great deal of research that can fall along these lines, such as \cite{huang2020conquering,FakeDomains20,WWWDiffusion20}. But it is not always clear to what degree other features are orthogonal to content, and explicit analysis in future work could be useful.

Our results on classifying tweet IDs show that depending on the collection strategy, the tweet date may be surprisingly informative for the labels. This means the distribution of data may not be a good match for real-world tasks. A simple random forest on the first few digits of the tweet ID can be used to detect if this issue is present in the data. We suggest this test be applied to future datasets to ensure a reasonable level of temporal randomization, or to find and report that it is not random and thus to facilitate designing algorithms and interpreting results in light of that. Similar techniques can be applied to data from other social networks.

It is also possible to time-randomize data after the primary collection. In this domain, the fake examples are typically the ones that are hard to find, as in most contexts the majority of tweets are real \cite{Grinberg374}. We can time-randomize without retrieving new fake tweets if we can retrieve additional real tweets (and tweets from the other classes, when working with more than real/fake) from approximately the same time as the fake ones. Then we can replace the original real tweets, yielding both fake and real tweets that are hard to distinguish using time. This may be facilitated by the new Twitter Academic API,\footnote{https://developer.twitter.com/en/solutions/academic-research} which gives more historical access than the normal API. Since Twitter16 is commonly used \cite{huang2020heterogeneous, RumorSleuth19, veyseh2019rumor}, has severe time-determinacy, and does not have too many tweets to deal with, future work to improve this dataset in particular could be worthwhile, though the keyword issues with fake tweets might be hard to fix.

\section{Conclusion}

Overall, our findings here highlight the need to combine both development of better algorithms and data science. To solve critical real-world problems like misinformation, we need better understandings of how different models and datasets compare and interact. Otherwise we risk creating sophisticated models that are beaten by brute force approaches - applying the biggest LM one can run - or models that work well on standard datasets but poorly in practice, e.g. by learning to predict the date or keywords like "Steve." The SOTA models we compare with in this work have more to offer than pure performance (for example, explainability), and there are other tasks like early detection that we have not examined here. But we suggest future work can benefit from adjusting and further considering how we frame these problems. This will help not only build higher metrics but also real-world solutions. 

Besides considerations for improving current lines of research, this paper suggests two other promising directions for future work:

\begin{itemize}
    \item Benchmarking standard misinformation datasets and publishing exact splits. We see, for example on PHEME, that there are many ways researchers have split the data, and benchmarks are lacking. This leads to difficulty for the research community in comparing results. Better benchmarks and the ability to compare exact splits would help mitigate this issue.
    \item New, thoroughly benchmarked and validated datasets. Although Twitter15 and Twitter16 have flaws highlighted in this paper, they also have excellent propagation information, which has likely motivated many of the approaches that leverage that to use those datasets. Improved datasets should lead to improved real-world results.
\end{itemize}

Finally, we hope to integrate these models with work on other features and data types, and produce thoroughly evaluated models that can combine the best of both recent language models and misinformation detection domain algorithms.



\bibliographystyle{ACM-Reference-Format}
\bibliography{main.bib}





\end{document}